\documentclass[a4paper,fleqn]{cas-dc}
\usepackage{float}
\usepackage{soul}
\usepackage[numbers]{natbib}
\usepackage{subfigure} 
\def\tsc#1{\csdef{#1}{\textsc{\lowercase{#1}}\xspace}}
\tsc{WGM}
\tsc{QE}
\tsc{EP}
\tsc{PMS}
\tsc{BEC}
\tsc{DE}

\begin{document}
\let\printorcid\relax
\let\WriteBookmarks\relax
\def\floatpagepagefraction{1}
\def\textpagefraction{.001}
\shorttitle{Leveraging social media news}
\shortauthors{Weiming Hu et~al.}

\title[mode = title]{
GasHisSDB: A New Gastric Histopathology Image Dataset for Computer Aided Diagnosis of Gastric Cancer
}  

\author[a]{Weiming Hu}[type=editor,style=chinese,auid=000,bioid=1,]
\author[a]{Chen Li}[style=chinese]
\cormark[1]
\ead{lichen201096@hotmail.com}
\author[b]{Xiaoyan Li}[style=chinese]
\author[a]{Md Mamunur Rahaman}
\author[c]{Jiquan Ma}
\author[b]{Yong Zhang}[style=chinese]
\author[a]{Haoyuan Chen}[style=chinese]
\author[a]{Wanli Liu}[style=chinese]
\author[a]{Changhao Sun}[style=chinese]
\author[d]{Yudong Yao}
\author[e]{Hongzan Sun}
\author[f]{Marcin Grzegorzek}

\address[a]{Microscopic Image and Medical Image Analysis Group, MBIE College, Northeastern University, 110169, Shenyang, PR China}
\address[b]{Department of Pathology, Cancer Hospital, China Medical University, Liaoning Cancer Hospital and Institute,
Shenyang 110042, PR China}
\address[c]{Department of Computer Science and Technology, Heilongjiang University, Harbin, Heilongjiang, 150080, China}
\address[d]{Department of Electrical and Computer Engineering, Stevens Institute of Technology, Hoboken, NJ 07030, USA}
\address[e]{'Department of Radiology, Shengjing Hospital, China Medical University, Shenyang, 110122, China}
\address[f]{Institute of Medical Informatics, University of Luebeck, Luebeck, Germany}
\cortext[cor1]{Corresponding author}

\begin{abstract}
Background and Objective: Gastric cancer has turned out to be the fifth most common cancer globally, and early detection of gastric cancer is essential to save lives. Histopathological examination of gastric cancer is the gold standard for the diagnosis of gastric cancer. However, computer-aided diagnostic techniques are challenging to evaluate due to the scarcity of publicly available gastric histopathology image datasets.\\
Methods: In this paper, a noble publicly available Gastric Histopathology Sub-size Image Database (GasHisSDB) is published to identify classifiers' performance. Specifically, two types of data are included: normal and abnormal, with a total of 245,196 tissue case images. In order to prove that the methods of different periods in the field of image classification have discrepancies on GasHisSDB, we select a variety of classifiers for evaluation. Seven classical machine learning classifiers, three Convolutional Neural Network classifiers, and a novel transformer-based classifier are selected for testing on image classification tasks.\\	
Results: This study performed extensive experiments using traditional machine learning and deep learning methods to prove that the methods of different periods have discrepancies on GasHisSDB. Traditional machine learning achieved the best accuracy rate of 86.08\% and a minimum of just 41.12\%.
The best accuracy of deep learning reached 96.47\% and the lowest was 86.21\%. Accuracy rates vary significantly across classifiers.\\
Conclusions: To the best of our knowledge, it is the first publicly available gastric cancer histopathology dataset containing a large number of images for weakly supervised learning. We believe that GasHisSDB can attract researchers to explore new algorithms for the automated diagnosis of gastric cancer, which can help physicians and patients in the clinical setting.
\end{abstract}



\maketitle

\section{Introduction}
\subsection{Research Background and Motivation}

Gastric cancer is a common type of malignant tumor 
with a relatively poor prognosis 
and presents a severe threat to global health~\cite{Wang-2019-tcsoc}.
According to the recent Global Cancer Statistics, 
gastric cancer is the fifth commonly diagnosed cancer~\cite{Sung-2021-gcs2g},
and accountable for 18.0\% of the total cancer deaths~\cite{Sung-2021-gcs2g}.
Histopathological examination of gastric cancer is 
a microscopic examination of paraffin sections 
made from tissues taken from suspected cancerous sections 
by experienced pathologists~\cite{Tom-2014-gceee}.
It is the gold standard for gastric cancer diagnosis and an essential requirement before treatment initiation~\cite{Wang-2019-tcsoc}.

Hematoxylin and Eosin (H\&E) staining is a commonly used staining method in paraffin section technology to demonstrate nucleus and cytoplasmic inclusions in clinical specimens and highlight the delicate structures of cells and tissues~\cite{Fischer-2008-haeso,Chan-2014-twcot,Cardiff-2014-mhaes}.
Firstly, the pathological segment is observed with naked eyes to observe whether the section is qualified or not to find the lesion roughly.
Then, the use of a low magnification microscope to observe and diagnose. Finally, if doctors want to observe the fine structure of the lesion more clearly, they can move it to the center of the visual fields, switch to a high-power microscope for the analysis~\cite{kumar-2017-rbpeb}.
It can be seen that there are the following problems 
in the process of observing H\&E staining sections 
of gastric cancer: the diagnosis results of medical doctors are subjective and hard to be quantitative described, their workload is heavy, and working hours are long, 
it is easy for doctors to omit section information, 
and it is inconvenient to use big data technology.
Therefore, there is an urgent need 
for more efficient solutions to related problems.

With the rapid development of computer vision technology, especially the emergence of medical image classification, it is possible to examine every electron microscopic photo quickly and efficiently~\cite{wang-2021-miras,tsekrekos-2019-htrgi}.
Thus, it brings an opportunity to solve the problems related to the diagnosis of gastric cancer~\cite{jin-2020-aiigc,mirniaharikandehei-2021-aarpa}.
Especially, image 
classification plays an important role 
in computer-aided diagnosis. In the field 
of distinguishing benign or malignant tumors, 
distinguishing the differentiation stage of tumors 
and distinguishing the subtype of cancer~\cite{le-2021-jcaol}, 
the results of image classification methods can be used as an important reference for clinicians in diagnostic practice.
Furthermore, with the development of medical image classification technology, the main purpose of this technology is to achieve high accuracy and have the high anti-interference ability
~\cite{Miranda-2016-asomi,affonso-2017-dlfbi}.
Although the mainstream trend is to scan 
the whole-slide images for analysis, the actual work often encounters the actual situation of computer performance shortage, 
where the whole-slide images are usually cropped 
into many sub-size images for analysis. 
To this end, we need a Sub-size Image Database 
to test the anti-interference ability 
of various medical image classification techniques
~\cite{kotadiya-2019-romic}.

This paper introduces a noble publicly available Gastric Histopathology Sub-size Image Database (GasHisSDB) , consisting of 245196 sub-size pathological images of gastric cancer and contains three sub-size labels.
Furthermore, each image in this database calculates three different features.
Moreover, the evaluation results of different classification schemes based on features and images, using machine learning and deep learning methods, are presented to show each classifier’s discrimination ability.
GasHisSDB is available at the URL:https://gitee.com/neuhwm/GasHisSDB.git

The main contributions of this paper are as follows:
\begin{itemize}
\itemsep=0pt
\item Development of a Gastric Histopathology Subsize Image Database based on open-source databases, in which several experts participated in the production of the database.
\item Demonstrate that GasHisSDB can be used to differentiate the performance of classifiers including traditional machine learning classifiers and deep learning classifiers.
\item This Gastric Histopathology Subsize Image Database is published as open source for non-commercial purposes.
\end{itemize}

\subsection{Related work}

In the field of classical machine learning methods for gastric cancer image classification, a commonly used open-source database is introduced in ~\cite{sharma-2015-amafc}.
This dataset consists of 11 Human Epidermal GrowthFactor Receptor 2 (HER2) and H\&E stained Whole-Slide Images (WSI) at a magnification of 40$\times$, annotated by ten pathologists, taken from surgical sections of patients with different gastric cancers~\cite{behrens-2015-rohsi}.
In classical machine learning classification studies, this dataset is often pre-processed, including operations such as semi-automatic registration and conversion of HER2 WSI to H\&E stained WSI~\cite{sharma-2017-acsof}.
Another commonly used open source database is the public database used in this paper~\cite{TP-2017-psioa}, which contains 560 abnormal images and 140 normal images. Studies often pre-process this database, including random cuts, geometric transformations~\cite{liu-2018-cogsb}.

Meanwhile, some gastric cancer image classification studies use proprietary databases, such as the study using stomach moleculer images taken from Pathology Department of the Medicine Faculty in the Fırat University~\cite{korkmaz-2018-comsi}.
In the study, the size of these images is 2592$\times$1944. 180 images are taken, of which 60 are normal, 60 are benign, and 60 are malignant.

The dataset in \cite{sharma-2015-amafc} is also public in the field of deep learning classification of gastric cancer images.
In \cite{sharma-2017-dcnnf}, this dataset is pre-processed with overlapping by a factor of 0.3, and affine transformations such as rotations with 10-degree intervals, reflection, rotation after reflection and shear by a factor of 0.1.
After a further 512$\times$512 cropping, this study yielded a total of 231000 initial data.
In the field of deep learning classification, studies are using the same base database as ~\cite{li-2018-dlbgc}. In this study, the image pre-processing operation is to crop the whole image into 224$\times$224 patches, obtaining 8992 abnormal patches and 14000 normal patches.

There are also deep learning studies of gastric cancer image classification using proprietary databases, as follows.
A new large Whole Slide Gastric Image dataset (WSGI) is constructed by collaborating with The Sixth Affiliated Hospital of Sun Yat-sen University in \cite{wang-2019-rrmdl}. 
The WSGI dataset consists of 608 complete slide images with a magnification of 40$\times$. The labels of WSGI include three categories: normal, dysplasia, and cancer.
The dataset used in \cite{zhu-2020-wbanf} contains 410 pathological images of gastric cancer of size 2048$\times$2048 and 210 images of normal tissue of the same size. This study rearrange and crop this dataset to 256$\times$256 yielding 5905 training images and 655 test images.
The deep learning model in the study of \cite{song-2020-cahds} is trained with 2123 pixel-level annotated  H\&E stained digital slides from 1500 patients, which include 958 surgical specimens (908 malignancies) and 542 biopsies (102 malignancies) with diverse tumor subtypes.
Testing is then performed on a multicenter dataset, which included 355 cases (595 slides) from Peking Union Medical College Hospital and 541 cases (987 slides) from Cancer Hospital of Chinese Academy of Medical Sciences.
In \cite{kosaraju-2020-dmrfd}, a histopathological image dataset is obtained from gastroscopic biopsy specimens of 94 cases at Gyeongsang National University Changwon Hospital (Changwon, Korea).
Then all 94 WSIs are manually categorized into four groups: 13 well-differentiated, 11 moderately-differentiated, 20 poorly-differentiated adenocarcinoma and 50 poorly cohesive carcinoma.
And poorly cohesive carcinoma include 20 signet-ring cell features and 30 normal gastric mucosa.

Table. \ref{tbl0} summarizes existing studies and their use dataset for gastric histopathological image classification methods.

The databases proposed in the above papers include large size images and are in small quantities. Data augmentation operations are often performed on the databases to expect better classification performance of the classifiers. The source database utilized in this paper also has the aforementioned problems. Therefore, this study makes full use of the already existing resources and complements the shortcomings of the original database functionality to make it more comprehensive.

\begin{table*}[]
\caption{Data usage in image classification of gastric cancer.}\label{tbl0}
\begin{tabular}{cccccc}
\hline
Aim                          & Year & Reference                                   & Team            & Categorization                                                                                                                                          & Amount      \\ \hline
                                                                             & 2015 & \cite{sharma-2015-amafc}   & Sharma et al.   & HER2+ tumor/HER2- tumor/Non-tumor                                                                                                                       & 11          \\
                                                                             & 2015 & \cite{sharma-2017-acsof}   & Sharma et al.   & HER2+ tumor/HER2- tumor/Non-tumor                                                                                                                       & 11          \\
                                                                             & 2017 & \cite{liu-2018-cogsb}      & Liu et al.      & Abnormal/Normal                                                                                                                                         & 560/140     \\
\multirow{-4}{*}{\begin{tabular}[c]{@{}c@{}}Machine\\ learning\end{tabular}} & 2018 & \cite{korkmaz-2018-comsi}  & Korkmaz et al.  & Normal/Benign/Malignant                                                                                                                                 & 60/60/60    \\ \hline
                                                                             & 2017 & \cite{sharma-2017-dcnnf}   & Sharma et al.   & HER2+ tumor/HER2- tumor/Non-tumor                                                                                                                       & 11          \\
                                                                             & 2018 & \cite{li-2018-dlbgc}       & Li et al.       & Abnormal/Normal                                                                                                                                         & 560/140     \\
                                                                             & 2019 & \cite{wang-2019-rrmdl}     & Wang et al.     & Normal/Dysplasia/Cancer                                                                                                                                 & 117/172/319 \\
                                                                             & 2020 & \cite{zhu-2020-wbanf}      & Zhu et al.      & Cancer/Normal                                                                                                                                           & 410/210     \\
                                                                             & 2020 & \cite{song-2020-cahds}     & Song et al.     & Malignancy/Normal                                                                                                                                       & 1010/490    \\
\multirow{-6}{*}{\begin{tabular}[c]{@{}c@{}}Deep\\ learning\end{tabular}}    & 2020 & \cite{kosaraju-2020-dmrfd} & Kosaraju et al. & \begin{tabular}[c]{@{}c@{}}Well-differentiated/Moderately-differentiated/poorly-\\ differentiated adenocarcinoma/poorly cohesive carcinoma\end{tabular} & 13/11/20/50 \\ \hline
\end{tabular}
\end{table*}

\subsection{Structure of this paper}

This section describes the background and motivation for the database preparation and does related work on the database of related research papers. Next, the methods of preparing the database and evaluating the database are described in detail, which includes the details of each item of the database. The third chapter shows the results of the evaluation database. Section 4 discusses the results of section 3. Finally a summary and future work is given.

\section{Method}
\subsection{Database preparation}
GasHisSDB contains 245,196 sub-size images in two categories, including 97,076 abnormal images and 148,120 normal images.
The details of the applied datasets are introduced as follows.

 \textbf{
Gastric Histopathology Sub-size Image Database:
}

\begin{figure*}
	\centering
		\includegraphics[scale=.7]{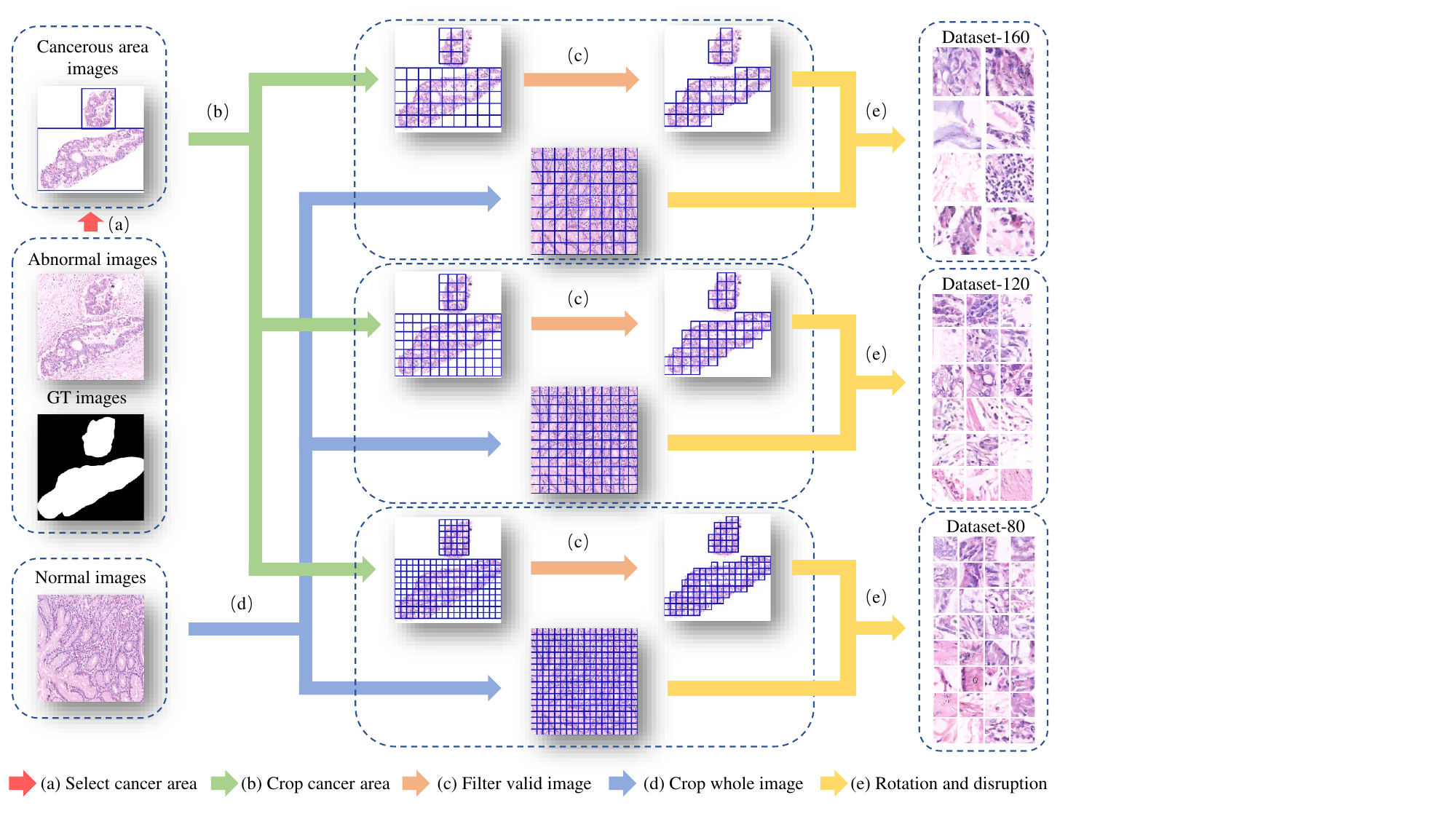}
		\caption{Data preparation workflow of GasHisSDB.}
	\label{FIG:1}
\end{figure*}

\begin{enumerate}
\itemsep=0pt
\item Data source: \\
Stage 1: Four pathologists from Longhua Hospital Shanghai University of Traditional Chinese Medicine provide 600 images of gastric cancer pathology images of size 2048×2048 and give tissue-level labels for a strongly supervised learning process~\cite{TP-2017-psioa,li-2018-dlbgc,sun-2020-ghisu}.\\
Stage 2: Based on stage 1, five biomedical researchers from Northeastern University prepare 245196 sub-sized gastric cancer pathology images for weakly supervised learning~\cite{zhang-2021-jfcag}, and two experienced pathologists from Liaoning Cancer Hospital and Institute perform the calibration.
\item Preparation rules: According to the existing method of making the database
~\cite{li-2019-chicu,mohamed-2021-fmrfa,potovcnik-2020-pdfvo},
the preparation rules of this database is as follows: \\
Rule 1: Three sizes 
($160\times160$, $120\times120$, $80\times80$ pixels)
of normal pathological sections are cropped directly. 
And, it is necessary to select the cancerous region as the region of interest when dealing with abnormal pathological sections.\\Rule 2:  The region of interest of the pathological section and the ground truth is cropped simultaneously. This work can be used to filter out images with very few cancerous areas. \\Rule 3: In order to reduce the correlation among sub-size images from the same original images, each one is rotated randomly, and the image order of the whole database is scrambled.\\ The data preparation workflow of GasHisSDB is shown in Figure. \ref{FIG:1}.
\item Staining: H\&E staining (See section 2.2.3 for details).
\item Magnification: 20$\times$.
\item Microscope: Nikon (Japan) and Olympus (Japan).
\item Acquisition software: NewUsbCamera.
\item Sub-database and image sizes: \\Sub-database A: 160$\times$160 pixels,
\\Sub-database B: 120$\times$120 pixels,
\\Sub-database C: 80$\times$80 pixels.
\item Scale: See Table. \ref{tbl1}.
\begin{table}[width=1.1\linewidth,cols=4,pos=h]
\caption{Dataset scale of GasHisSDB.}\label{tbl1}
\begin{tabular*}{\tblwidth}{@{} LLLLLLLL@{} }
\toprule
Sub-database name & Cropping size    & Abnormal & Normal  \\
\midrule
Sub-database A    & $160\times160$ pixels           & 13,124   & 20,160  \\
Sub-database B    & $120\times120$ pixels           & 24,801   & 40,460  \\
Sub-database C    & $80\times80$ pixels             & 59,151   & 87,500  \\
\bottomrule
Total             &                  & 97,076   & 148,120 \\
\bottomrule
\end{tabular*}
\end{table}
\item Image format:  "*.png".
\item Image types:\\
Normal: No cancerous cells appeared in the section (See section 2.2.1 for details);
Abnormal: Cancerous cells appear in this section (See section 2.2.2 for details).
\item The research and preparation of GasHisSDB is approved by the Ethical Committee at Northeastern University.
\end{enumerate}

\subsection{Database description}
\subsubsection{Normal image}
Every normal image does not contain cancerous region. 
Each cell has no or very small atypia (Figure. \ref{FIG:2} (a)). 
Moreover, the nuclei of the cells in the image 
have almost no mitosis and are regularly arranged in one layer
(Figure. \ref{FIG:2} (b)).
Therefore, when observed under the optic microscope, 
if no cancelation of any cells and tissues is observed, 
and the characteristics of the normal image are satisfied, 
it can be judged that this is a normal image~\cite{weiss-2007-eawst}.
In making normal image data set, we directly crop the whole image because of the characteristics of normal images.
The resulting images are shown in (a) (b) (c) 
in Figure. \ref{FIG:3}.

\subsubsection{Abnormal image}
\begin{figure}
	\centering
		\includegraphics[scale=.3]{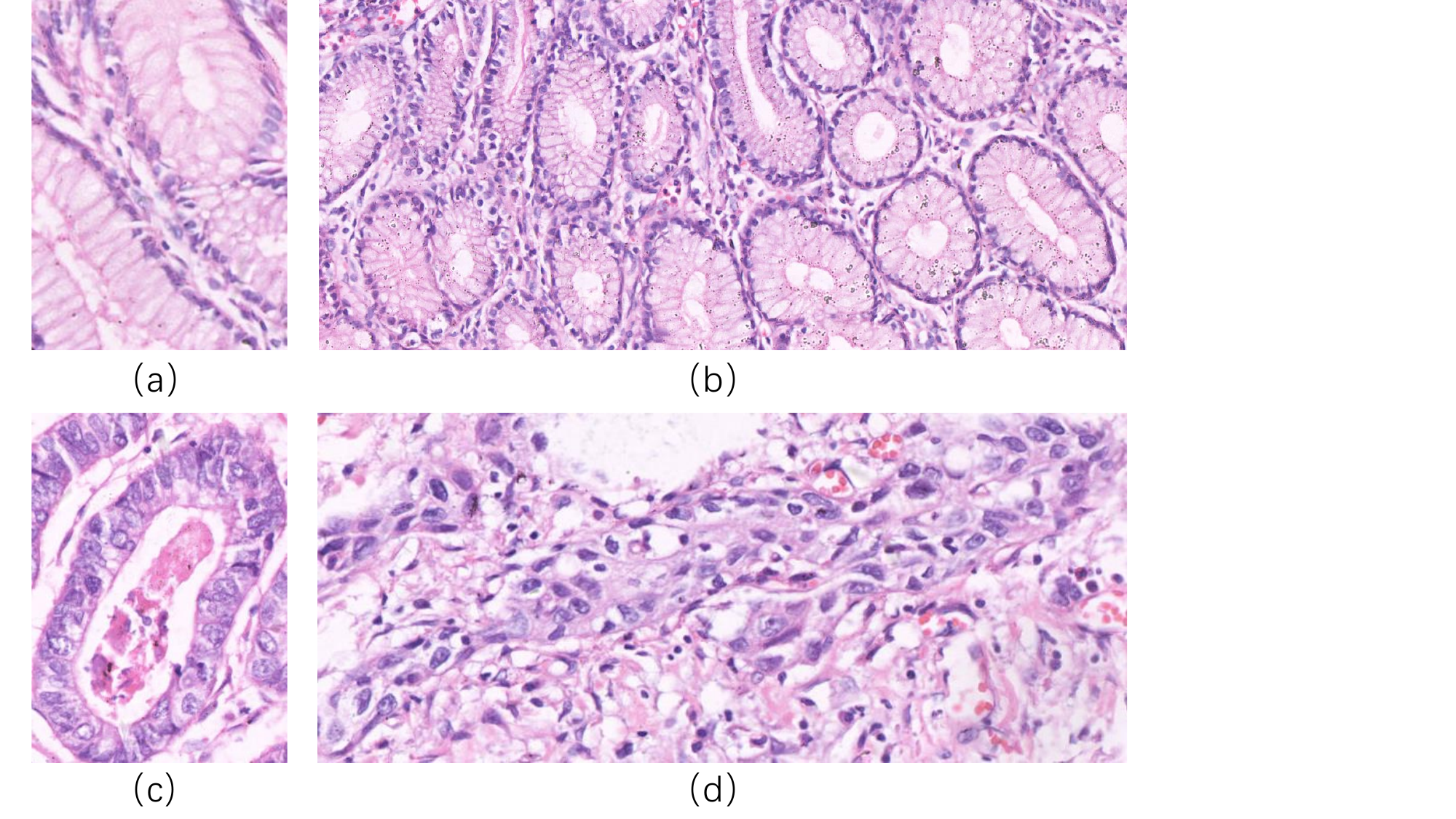}
		\caption{Example of Pathological images:
		(a)(b) Normal pathological images,
		(c) Non-invasive abnormal pathological image,
		(d) Invasive abnormal pathological image.}
	\label{FIG:2}
\end{figure}

\begin{figure}
	\centering
		\includegraphics[scale=.4]{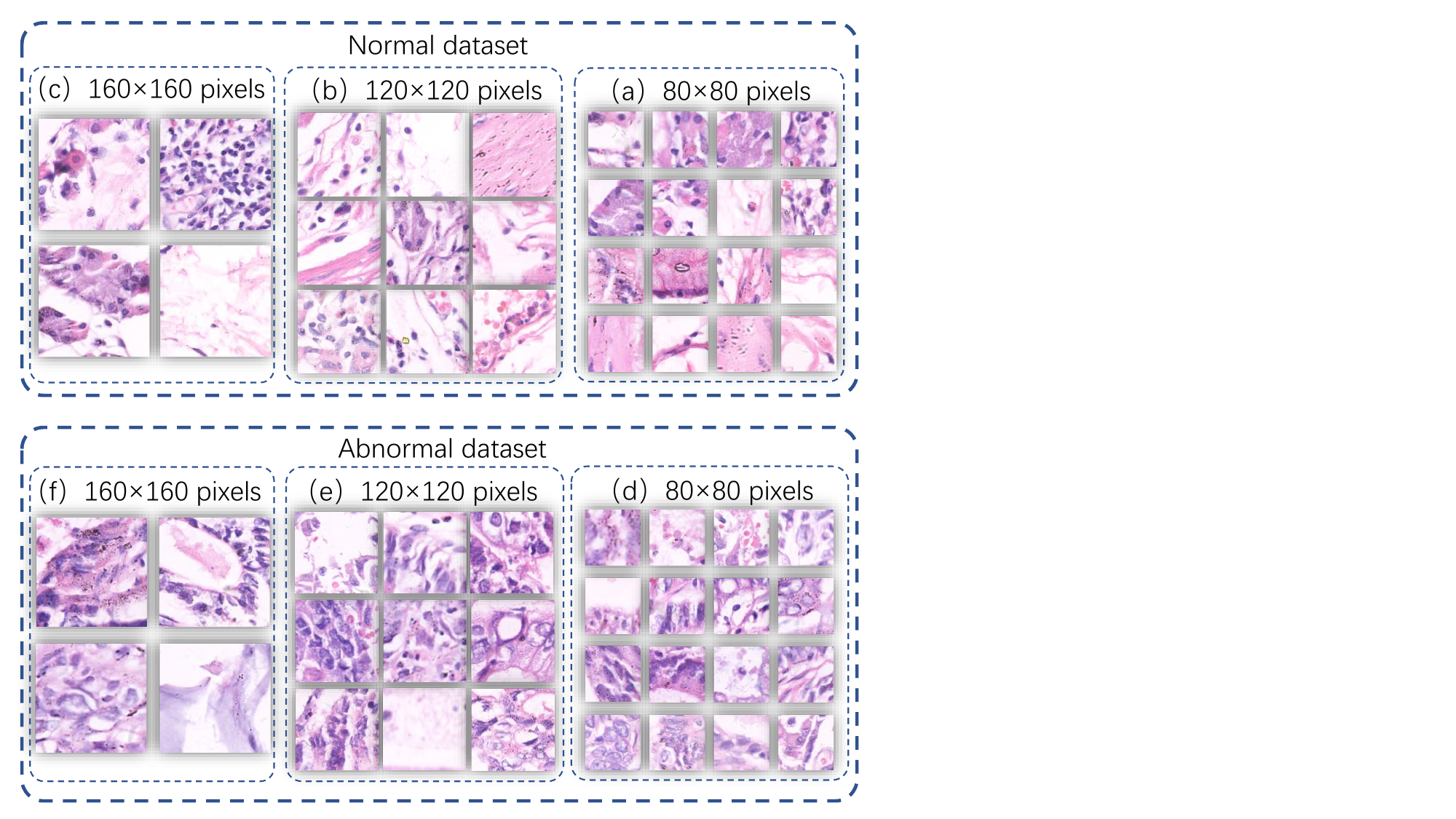}
		\caption{Example of GasHisSDB.}
	\label{FIG:3}
\end{figure}

Each abnormal image contains gastric cancer.
The general shape of gastric cancer is mostly ulcer type. As the disease progresses, cancer nests grow infiltrating from the mucosal layer to the muscle layer and serosal layer.
It has a hard texture, 
and the section is often gray-white. When observing under a microscope, the cancer cells can 
be arranged in a nest, acinar, tubular, or cord shape, 
and the boundary with the stroma is generally clear
(Figure. \ref{FIG:2} (c)). 
However, when the cancer cells infiltrate into the stroma, 
the boundary between them is not clear
(Figure. \ref{FIG:2} (d)).
Based on the above facts, when the cells are observed to form gland or adenoid structures with uneven size, different shapes, and irregular arrangement, it can be judged that the pathological image is abnormal.
In the abnormal images, the cancerous cells are often irregularly arranged in many layers, and the nuclei have different sizes and division phenomena~\cite{Wang-2019-tcsoc,japanese-2011-jcogc,takahashi-2013-gccso}.
In the process of making the abnormal image dataset, 
we crop every cancerous region selected according to 
the original ground truth (GT) images.
Then, the cropped images are filtered based on the cancerous area (usually 50\%) in the images. The resulted images are shown in Figure. \ref{FIG:3} - (d)(e)(f).

\subsubsection{Staining method}
H\&E staining method is one of the commonly used staining methods in paraffin section technology and is often the gold standard. The hematoxylin is alkaline and stains cell nuclei a purplish blue, and eosin is acidic and stains the extracellular matrix and cytoplasm pink, with other structures taking on different shades, hues, and combinations 
of these colors~\cite{Chan-2014-twcot,stevens-1990-th}. Therefore, pathologists can easily distinguish the nuclear and cytoplasmic parts of cells. In addition, the overall staining pattern of staining shows the overall layout and distribution of the cells and provides a general overview of the structure of the tissue sample~\cite{wittekind-2003-tsfrd}. H\&E staining method is the most basic and most widely used technical method in histology, embryology, pathology teaching and scientific research.

The pathological image after he staining has more pink and white areas in the normal image. Compared with the normal image, in the abnormal image, the purplish blue area is more distributed and messy.

\subsection{Methods of feature extraction}
Extracting various virtual features of a database is the preparation for classification using a machine learning classifier.
In this paper, we use three methods to extract the visual features of the database, including color histogram,  local binary pattern (LBP), and gray level co-occurrence matrix (GLCM) features.

\subsubsection{Color histogram}
Color histogram is the most commonly used method 
to describe the color characteristics of images. It can simply represent 
the global distribution of colors in an image, 
i.e., the proportion of different colors in the whole image. 
It is especially suitable for describing images that 
are difficult to segment automatically and images that do not 
need to consider the spatial location of objects~\cite{gevers-2006-cfd,van-2006-clfe}. 
Its advantage is that it is not affected by image rotation, shift changes, and further normalization by image scale changes.
The disadvantage is that it cannot describe the local distribution of colors in the image, each color's spatial location, and specific objects.
\subsubsection{Texture features}

The texture is a visual feature that reflects homogeneity in an 
image. It reflects the organization and arrangement of surface 
structures with slow or periodic changes on the surface of an object. 
It is not based on the characteristics of pixels but needs to carry 
out the statistical calculation on the region containing multiple 
pixels~\cite{sharma-2001-eotmf}. The texture is represented by the gray distribution of pixels and their surrounding spatial neighbor, local texture information. Besides, global texture 
information is the degree of repetition of local texture information.
This paper uses two methods to describe texture features of 
GasHisSDB, which are Local Binary Patterns (LBP) and Gray-level Co-occurrence Matrix (GLCM).

LBP is an effective texture description operator that measures and 
extracts textural information local to an image with significant 
advantages such as grey level invariance and rotational invariance, 
and the feature is easy to compute. The value of LBP for each pixel 
is eight values within its eight-neighbor compared 
to it's pixel. If 
the pixel value in the eight-neighbor is greater than or equal to the 
central pixel value, the position is labeled by 1, and 0 for otherwise, 
aligning them clockwise after eight comparisons yield a binary number 
of eight lengths, yielding LBP values~\cite{zhou-2013-ltfbo}.

The GLCM is defined by the joint probability density 
of the pixels at two locations. It not only reflects the distribution 
of brightness but also reflects the distribution of location between 
pixels with the same or near brightness. It is a second-order 
statistical feature about the change of image brightness. GLCM is the 
basis for defining a set of texture features~
\cite{mohanaiah-2013-itfeu,qunqun-2013-eocit,ali-2018-cagad}. To more intuitively 
describe the texture state with a symbiotic matrix, some parameters 
reflecting the state of the matrix are derived from the symbiotic 
matrix, typically the following:                                    

\begin{enumerate}
\itemsep=0pt
\item Contrast: Reflects the sharpness of the image 
and the depth of texture grooves.
\item Correlation: It measures the degree of 
similarity of spatial grayscale symbiosis matrix elements in a row or 
column directions.
\item Energy: It is the sum of squares of gray 
symbiosis matrix elements, so it is also called energy, which 
reflects the uniformity of gray distribution and texture thickness of 
the image.
\item Homogeneity: Returns a measure of the 
compactness of the diagonal distribution of elements in GLCM.
\end{enumerate}

\subsection{Methods of classification}

After the steps of feature extraction, two classical machine learning methods 
are used to classify GasHisSDB, including Random Forest 
(RF) and linear Support Vector Machine (linear SVM). Furthermore, three 
classical or novel deep learning methods are used to classify 
GasHisSDB, including VGG16, ResNet50 and ViT.

\subsubsection{Classical machine learning methods}
The machine learning method for classification interpreters
whether the image is normal or not by its visual features.
RF is a parallel ensemble learning method and an extended variant of Bagging. 
RF is based on a decision tree learner, which adds random attribute selection 
to the training process of the 
decision tree~\cite{liaw-2002-carbr}.
SVMs are classified into linear and non-linear according to the kernel 
functions. Linear SVM maps training examples to points in space to maximize 
the gap between the two categories. Then, the new examples are mapped to the 
same space and predicted to belong to a category based on which side of the 
gap they fall on.

\subsubsection{Deep learning methods}
The concept of deep learning originates from the research of Artificial Neural Network~
\cite{hassoun-1995-foann}, where a 
multi-layer perceptron with multiple hidden layers is a deep learning 
structure. Deep learning forms a more abstract high-level representation 
attribute category or feature by combining low-level features to discover 
distributed feature representations of data~\cite{wang-2020-anact}.

In 2014, the Visual Geometry Group and Google DeepMind developed a new deep 
convolutional neural network: VGG~\cite{simonyan-2014-vdcnf}.
VGG is a \emph{Convolutional Neural Network} (CNN) improved by AlexNet. 
Several forms of VGG models are released, and the most commonly used in the 
field of image classification is VGG16 one.
In VGG16, three 3$\times$3 convolution kernels are used instead of 7$\times$7 
convolution kernels, and two 3$\times$3 convolution kernels are used instead 
of 5$\times$5 convolution kernels. The main purpose of this structure is to 
ensure the same perceptual field conditions, enhance the depth of the 
network, and to a certain extent enhance the effect of the neural network.

In 2016, to solve the difficulty of training deep networks due to the disappearance of gradients, Kaiming He et al. proposed  various forms of ResNet~\cite{he-2016-drlfi}. The most commonly used in the field of image classification is ResNet50. The ResNet50 team separately constructed a ResNet50 building block with "Shortcut Connection" and a down-sampling ResNet50 building block. A 1$\times$1 convolution operation is added to the main branch of the regional down-sampling building block.

In 2020, Alexey Dosovitskiy et al. proposed the ViT
model~\cite{dosovitskiy-2020-aiiww} by using the transformer, 
which is very effective in the field of natural language processing. It also 
provides good results in the image classification domain and reduces the 
dependence on CNNs.
ViT crops an image into small chunks and provides a sequence of linear embeddings of these chunks as input to the Transformer and trains the model in a supervised manner for image classification.

\section{Result}

\subsection{Results of classical machine learning methods}
In this paper, various classification experiments are conducted on the GasHisSDB database ($160\times160$ pixels, $120\times120$ pixels and $80\times80$ pixels) to demonstrate that GasHisSDB can be used to discriminate the performance of classifiers. Table. \ref{tbl2} show the comparison results of the classical machine learning methods. All the classification comparison experiments use the same parameters. The number of trees in the RF is set to 10. The SVM kernel function uses a linear kernel.

\begin{table*}[width=2\linewidth,cols=11,pos=h]
\caption{ Classification results of three image features using different classifiers in three  sub-databases of GasHisSDB (In [\%]). The bold text in the table indicates the highest value of the classification result for the same sub-database.}\label{tbl2}
\resizebox{\textwidth}{40mm}{
\begin{tabular}{cclrrrrrrrrrrrrr}
\hline
\multirow{2}{*}{\textbf{Sub-database}}                                                            & \multirow{2}{*}{\textbf{Freatures}} & \multicolumn{1}{c}{\multirow{2}{*}{\textbf{Methods}}} & \multicolumn{1}{c}{\multirow{2}{*}{\textbf{Acc}}} & \multicolumn{1}{c}{\textbf{}} & \multicolumn{1}{l}{\textbf{}} & \multicolumn{4}{c}{\textbf{Abnormal}}                                                                                                                           & \multicolumn{1}{c}{\textbf{}} & \multicolumn{1}{l}{\textbf{}} & \multicolumn{4}{c}{\textbf{Normal}}                                                                                                                             \\ \cline{7-10} \cline{13-16} 
                                                                                                  &                                     & \multicolumn{1}{c}{}                                  & \multicolumn{1}{c}{}                              & \multicolumn{1}{c}{\textbf{}} & \multicolumn{1}{l}{\textbf{}} & \multicolumn{1}{l}{\textbf{Precision}} & \multicolumn{1}{l}{\textbf{Recall}} & \multicolumn{1}{l}{\textbf{Specificity}} & \multicolumn{1}{l}{\textbf{F1-score}} & \multicolumn{1}{l}{\textbf{}} & \multicolumn{1}{l}{\textbf{}} & \multicolumn{1}{l}{\textbf{Precision}} & \multicolumn{1}{l}{\textbf{Recall}} & \multicolumn{1}{l}{\textbf{Specificity}} & \multicolumn{1}{l}{\textbf{F1-score}} \\ \hline
\multirow{6}{*}{\begin{tabular}[c]{@{}c@{}}Sub-database A\\ (160$\times$160 pixels)\end{tabular}} & \multirow{2}{*}{Color   histogram}  & RF                                                    & \textbf{85.99}                                    &                               &                               & 81.65                                  & 87.83                               & 84.55                                    & 84.63                                 &                               &                               & 89.88                                  & 84.55                               & 87.83                                    & 87.13                                 \\
                                                                                                  &                                     & linear SVM                                            & 41.12                                             &                               &                               & 33.92                                  & 35.96                               & 45.16                                    & 34.91                                 &                               &                               & 47.40                                  & 45.16                               & 35.96                                    & 46.25                                 \\
                                                                                                  & \multirow{2}{*}{LBP}                & RF                                                    & 70.27                                             &                               &                               & 62.16                                  & 62.84                               & 75.10                                    & 62.50                                 &                               &                               & 75.64                                  & 75.10                               & 62.84                                    & 75.37                                 \\
                                                                                                  &                                     & linear SVM                                            & 48.17                                             &                               &                               & 36.83                                  & 44.02                               & 50.87                                    & 40.10                                 &                               &                               & 58.27                                  & 50.87                               & 44.02                                    & 54.32                                 \\
                                                                                                  & \multirow{2}{*}{GLCM}               & RF                                                    & 71.39                                             &                               &                               & 63.16                                  & 65.85                               & 75.00                                    & 64.48                                 &                               &                               & 77.14                                  & 75.00                               & 65.85                                    & 76.06                                 \\
                                                                                                  &                                     & linear SVM                                            & 66.50                                             &                               &                               & 55.89                                  & 71.27                               & 63.39                                    & 62.65                                 &                               &                               & 77.22                                  & 63.39                               & 71.27                                    & 69.63                                 \\ \hline
\multirow{6}{*}{\begin{tabular}[c]{@{}c@{}}Sub-database B\\ (120$\times$120 pixels)\end{tabular}} & \multirow{2}{*}{Color histogram}    & RF                                                    & \textbf{86.08}                                    &                               &                               & 80.36                                  & 83.87                               & 87.43                                    & 82.08                                 &                               &                               & 89.84                                  & 87.43                               & 83.87                                    & 88.62                                 \\
                                                                                                  &                                     & linear SVM                                            & 46.28                                             &                               &                               & 39.48                                  & 77.62                               & 27.06                                    & 52.34                                 &                               &                               & 66.36                                  & 27.06                               & 77.62                                    & 38.45                                 \\
                                                                                                  & \multirow{2}{*}{LBP}                & RF                                                    & 70.13                                             &                               &                               & 60.88                                  & 59.90                               & 76.41                                    & 60.39                                 &                               &                               & 75.66                                  & 76.41                               & 59.90                                    & 76.03                                 \\
                                                                                                  &                                     & linear SVM                                            & 46.21                                             &                               &                               & 29.70                                  & 30.40                               & 55.89                                    & 30.05                                 &                               &                               & 56.71                                  & 55.89                               & 30.40                                    & 56.30                                 \\
                                                                                                  & \multirow{2}{*}{GLCM}               & RF                                                    & 71.15                                             &                               &                               & 61.42                                  & 64.72                               & 75.09                                    & 63.03                                 &                               &                               & 77.64                                  & 75.09                               & 64.72                                    & 76.34                                 \\
                                                                                                  &                                     & linear SVM                                            & 66.66                                             &                               &                               & 55.02                                  & 67.30                               & 66.28                                    & 60.54                                 &                               &                               & 76.78                                  & 66.28                               & 67.30                                    & 71.14                                 \\ \hline
\multirow{6}{*}{\begin{tabular}[c]{@{}c@{}}Sub-database C\\ (80$\times$80 pixels)\end{tabular}}   & \multirow{2}{*}{Color histogram}    & RF                                                    & \textbf{83.27}                                    &                               &                               & 77.14                                  & 83.15                               & 83.34                                    & 80.03                                 &                               &                               & 87.98                                  & 83.34                               & 83.15                                    & 85.60                                 \\
                                                                                                  &                                     & linear SVM                                            & 60.81                                             &                               &                               & 50.86                                  & 83.58                               & 45.41                                    & 63.24                                 &                               &                               & 80.36                                  & 45.41                               & 83.58                                    & 58.03                                 \\
                                                                                                  & \multirow{2}{*}{LBP}                & RF                                                    & 68.16                                             &                               &                               & 60.13                                  & 62.49                               & 71.98                                    & 61.29                                 &                               &                               & 73.95                                  & 71.98                               & 62.49                                    & 72.95                                 \\
                                                                                                  &                                     & linear SVM                                            & 43.10                                             &                               &                               & 27.68                                  & 25.48                               & 55.01                                    & 26.53                                 &                               &                               & 52.20                                  & 55.01                               & 25.48                                    & 53.56                                 \\
                                                                                                  & \multirow{2}{*}{GLCM}               & RF                                                    & 68.39                                             &                               &                               & 60.13                                  & 64.23                               & 71.21                                    & 62.11                                 &                               &                               & 74.65                                  & 71.21                               & 64.23                                    & 72.89                                 \\
                                                                                                  &                                     & linear SVM                                            & 66.82                                             &                               &                               & 57.14                                  & 71.04                               & 63.97                                    & 63.33                                 &                               &                               & 76.57                                  & 63.97                               & 71.04                                    & 69.71                                 \\ \hline
\end{tabular}}
\end{table*}

\subsection{Results of deep learning methods}

\begin{table*}[width=2\linewidth,cols=11,pos=h]
\caption{Classification results of four deep learning classifiers on GasHisSDB (In [\%]). The bold text in the table indicates the maximum value or the best index of the classification results of different categories.}\label{tbl5}
\resizebox{\textwidth}{50mm}{
\begin{tabular}{ccccccclllll}
\hline
\multicolumn{1}{l}{\textbf{Sub-database}}                                                         & \multicolumn{1}{l}{\textbf{Model}} & \multicolumn{1}{l}{\textbf{Quantity of epoch}} & \multicolumn{1}{l}{\textbf{Model size}} & \multicolumn{1}{l}{\textbf{best eopch}} & \multicolumn{1}{l}{\textbf{training time}} & \multicolumn{1}{l}{\textbf{Acc}} & \textbf{Category} & \textbf{Precision} & \textbf{Recall} & \textbf{Specificity} & \textbf{F1-score} \\ \hline
\multirow{8}{*}{\begin{tabular}[c]{@{}c@{}}Sub-database A\\ (160$\times$160 pixels)\end{tabular}} & \multirow{2}{*}{VGG16}             & \multirow{2}{*}{100}                           & \multirow{2}{*}{268.16}                 & \multirow{2}{*}{100}                    & \multirow{2}{*}{13873}                     & \multirow{2}{*}{95.90}           & Abnormal          & 93.8               & 96.0            & 95.9                 & 94.9              \\
                                                                                                  &                                    &                                                &                                         &                                         &                                            &                                  & Normal            & 97.3               & 95.9            & 96.0                 & 96.6              \\
                                                                                                  & \multirow{2}{*}{ResNet50}          & \multirow{2}{*}{100}                           & \multirow{2}{*}{83.12}                  & \multirow{2}{*}{84}                     & \multirow{2}{*}{10023}                     & \multirow{2}{*}{\textbf{96.09}}  & Abnormal          & 94.6               & 95.6            & 96.4                 & 95.1              \\
                                                                                                  &                                    &                                                &                                         &                                         &                                            &                                  & Normal            & 97.1               & 96.4            & 95.6                 & 96.7              \\
                                                                                                  & \multirow{4}{*}{ViT}               & \multirow{2}{*}{100}                           & \multirow{2}{*}{31.17}                  & \multirow{2}{*}{97}                     & \multirow{2}{*}{2587}                      & \multirow{2}{*}{86.21}           & Abnormal          & 83.8               & 80.6            & 89.9                 & 82.2              \\
                                                                                                  &                                    &                                                &                                         &                                         &                                            &                                  & Normal            & 87.7               & 89.9            & 80.6                 & 88.8              \\
                                                                                                  &                                    & \multirow{2}{*}{400}                           & \multirow{2}{*}{31.17}                  & \multirow{2}{*}{399}                    & \multirow{2}{*}{10014}                     & \multirow{2}{*}{92.23}           & Abnormal          & 92.1               & 87.8            & 95.1                 & 89.9              \\
                                                                                                  &                                    &                                                &                                         &                                         &                                            &                                  & Normal            & 92.3               & 95.1            & 87.8                 & 93.7              \\ \hline
\multirow{8}{*}{\begin{tabular}[c]{@{}c@{}}Sub-database B\\ (120$\times$120 pixels)\end{tabular}} & \multirow{2}{*}{VGG16}             & \multirow{2}{*}{100}                           & \multirow{2}{*}{268.16}                 & \multirow{2}{*}{100}                    & \multirow{2}{*}{26105}                     & \multirow{2}{*}{\textbf{96.47}}  & Abnormal          & 96.7               & 94.0            & 98.0                 & 95.3              \\
                                                                                                  &                                    &                                                &                                         &                                         &                                            &                                  & Normal            & 96.4               & 98.0            & 94.0                 & 97.2              \\
                                                                                                  & \multirow{2}{*}{ResNet50}          & \multirow{2}{*}{100}                           & \multirow{2}{*}{83.12}                  & \multirow{2}{*}{94}                     & \multirow{2}{*}{19087}                     & \multirow{2}{*}{95.94}           & Abnormal          & 96.2               & 93.0            & 97.8                 & 94.6              \\
                                                                                                  &                                    &                                                &                                         &                                         &                                            &                                  & Normal            & 95.8               & 97.8            & 93.0                 & 96.8              \\
                                                                                                  & \multirow{4}{*}{ViT}               & \multirow{2}{*}{100}                           & \multirow{2}{*}{31.17}                  & \multirow{2}{*}{100}                    & \multirow{2}{*}{4077}                      & \multirow{2}{*}{89.44}           & Abnormal          & 87.0               & 84.9            & 92.2                 & 85.9              \\
                                                                                                  &                                    &                                                &                                         &                                         &                                            &                                  & Normal            & 90.9               & 92.2            & 84.9                 & 91.5              \\
                                                                                                  &                                    & \multirow{2}{*}{500}                           & \multirow{2}{*}{31.17}                  & \multirow{2}{*}{496}                    & \multirow{2}{*}{20410}                     & \multirow{2}{*}{94.59}           & Abnormal          & 93.5               & 93.4            & 95.3                 & 93.2              \\
                                                                                                  &                                    &                                                &                                         &                                         &                                            &                                  & Normal            & 95.4               & 95.9            & 92.5                 & 95.6              \\ \hline
\multirow{8}{*}{\begin{tabular}[c]{@{}c@{}}Sub-database C\\ (80$\times$80 pixels)\end{tabular}}   & \multirow{2}{*}{VGG16}             & \multirow{2}{*}{100}                           & \multirow{2}{*}{268.16}                 & \multirow{2}{*}{90}                     & \multirow{2}{*}{62152}                     & \multirow{2}{*}{\textbf{96.12}}  & Abnormal          & 94.2               & 96.3            & 96.0                 & 95.2              \\
                                                                                                  &                                    &                                                &                                         &                                         &                                            &                                  & Normal            & 97.4               & 96.0            & 96.3                 & 96.7              \\
                                                                                                  & \multirow{2}{*}{ResNet50}          & \multirow{2}{*}{100}                           & \multirow{2}{*}{83.12}                  & \multirow{2}{*}{97}                     & \multirow{2}{*}{41992}                     & \multirow{2}{*}{96.09}           & Abnormal          & 96.2               & 94.0            & 97.5                 & 95.1              \\
                                                                                                  &                                    &                                                &                                         &                                         &                                            &                                  & Normal            & 96.0               & 97.5            & 94.0                 & 96.7              \\
                                                                                                  & \multirow{4}{*}{ViT}               & \multirow{2}{*}{100}                           & \multirow{2}{*}{31.17}                  & \multirow{2}{*}{89}                     & \multirow{2}{*}{8247}                      & \multirow{2}{*}{90.23}           & Abnormal          & 86.3               & 90.1            & 90.3                 & 88.2              \\
                                                                                                  &                                    &                                                &                                         &                                         &                                            &                                  & Normal            & 93.1               & 90.3            & 90.1                 & 91.7              \\
                                                                                                  &                                    & \multirow{2}{*}{500}                           & \multirow{2}{*}{31.17}                  & \multirow{2}{*}{496}                    & \multirow{2}{*}{41135}                     & \multirow{2}{*}{94.57}           & Abnormal          & 93.1               & 93.4            & 95.3                 & 93.2              \\
                                                                                                  &                                    &                                                &                                         &                                         &                                            &                                  & Normal            & 95.6               & 95.3            & 93.4                 & 95.4              \\ \hline
\end{tabular}}
\end{table*}

In this part, classical and novel deep learning methods are used to classify the Sub-database A, Sub-database B and Sub-database C of GasHisSDB.
In a series of comparative experiments, the ratio of the training, validation and test sets to the three Sub-databases is split 4:4:2.
Each model uses a learning rate of 0.00002, the batch size is set to 32, and the experiment is performed for 100 epochs, to observe the classification results of this database on different models.
The results of the comparative experiment are shown in Table. \ref{tbl5}.

\subsection{Additional experiment}

Since it is observed that the ViT did not fully converge in 100 epochs, a series of additional experiments are performed in this section.
The experimental results are shown in the last item of each Sub-database in Table. \ref{tbl5}.
In the follow-up additional experiment of Sub-database A, under the same parameter conditions, the training time is controlled to be similar to that of ResNet with running 100 epochs, so ViT needed to run 400 epochs.

\section{Discussion}

\subsection{Classical machine learning results discussion}

Sub-database A has the largest sample size and the smallest number of samples.
The color histogram is the extracted feature with the most feature items.
According to Table. \ref{tbl2}, the color histogram obtains the best classification accuracy in the classification results of different classifiers using three different features on Sub-database A in the RF, which is 85.99\%. 
However, the linear SVM classifier performs poorly on color histogram features, less than 50\%. 
As a comparison, the classification effect on LBP does not perform as well as the color histogram, and the highest classification effect on RF is only 70.27\%. And the classification effect of linear support vector machine on LBP is also less than 50\%.
GLCM contains the least amount of data among all features, with only four items per image.
The classification results of RF on GLCM are similar to those on LBP, but the linear SVM has a significant improvement in classification accuracy reaching 66.50\% on GLCM.

Sub-database B has more balanced sample size and sample number.
Overall, there was only a slight change in the classification accuracy of Sub-database B compared to the classification results of Sub-database A.
On Sub-database B, the best classification accuracy remained on the color histogram using RF classification, with a result of 86.08\%.
The RF is greater than 70\% on all three features and performs consistently. Another classical machine learning classifier, linear SVM, achieves 66.66\% on Sub-database B only on GLCM and performs poorly on the remaining features.

Sub-database C has the largest number of samples.
In Sub-database C, RF changed its classification result on the color histogram to 83.27\%, but still performed the best.
Compared with the other two Sub-databases, the linear SVM classification results differed more reaching 60.81\%.
The classification accuracy of RF decreased slightly on LBP and GLCM, both below 70\%.
Linear SVM on LBP still underperforms.
The linear SVM still underperforms on LBP, but performs the same on GLCM as on the other two sub-datasets.

\subsection{Deep learning results discussion}

In general, deep learning models are far superior to classical machine learning methods, and even the lowest ViT accuracy is still higher than the highest accuracy of classical machine learning methods. 
In Sub-database A, The accuracy of the VGG model is more than 95\%, but it has the longest training time and the model size is much larger than other deep learning models.
The accuracy of ResNet50 is slightly higher than that of VGG16, which is the highest value among all methods reaching 96.09\%. Both the model size and training time of ResNet50 are better than VGG16.
ViT is the latest model based on transformer structure. Its accuracy is at least 86.21\%, but it is still higher than the classification accuracy of RF on the color histogram. Importantly, compared to ResNet, ViT achieves such accuracy in only 1/4 of the training time and 1/3 of the model size. Moreover, the accuracy curve still has an increasing trend and the loss function still has not fully converged.

Because there are more training samples, VGG16 is the classifier with the highest accuracy in the classification results of the deep learning method of Sub-database B, which is 96.47\%.
The corresponding training time has also become twice that of Sub-database A.
The accuracy of ResNet50 almost caught up with that of VGG16 with 95.94\% accuracy. The accuracy of ViT also improved somewhat with the increase in the amount of training data, reaching 89.44\%.

Sub-database C is the Sub-database with the largest number of samples.
The accuracy of the four classifiers has only slightly changed. The classification model with the highest accuracy rate is still 96.12\% for VGG16.
The lowest accuracy rate is still the ViT model with the least training time, which is 90.23\%.
However, the training time of ViT in the Sub-database C has almost become 13.26\% of the training time of VGG16 with the highest accuracy.

\subsection{Additional experimental results discussion}

This experiment achieved an accuracy of 92.23\%.
Moreover, as the amount of data increases, when the control training time is the same as RseNet50, the accuracy of the ViT model with Sub-database B and Sub-database C is increased to 94.59\% and 94.57\%.
These image classification results have reached the general level of medical image classification, and the ViT model is excellent in terms of model size.

\subsection{Overall performance evaluation}

This chapter compares the different classifier's classification results from the Linear Regression to Visual Transformer on the Sub-database A, Sub-database B and Sub-databases C of the GasHisSDB.
Each method has a completely different classification performance on GasHisSDB. Classical machine learning methods have strict theoretical foundations, simple ideas, and excellent performance in some specific features and algorithms. However, deep learning methods are still far ahead of classical machine learning methods in terms of image classification accuracy and experimental workload.
As a sub-size image database, GasHisSDB can be proved to have distinguishing in a variety of methods.

\section{Conclusion and futures works}
In this paper, a sub-size gastric histopathological image database is developed, namely GasHisSDB.
GasHisSDB have three Sub-databases, $160\times160$ pixels Sub-database, $120\times120$ pixels Sub-database and $80\times80$ pixels Sub-database. Each Sub-database contains two folders of normal images and abnormal images.
Each folder contains cropped images that have been renamed and shuffled.
GasHisSDB has the function of discriminating the performance of classifiers.
This paper is divided into the following two parts for testing.
For classical machine learning methods, this paper extracts five different features.
Then test the classification performance of seven different classification methods on three sub-databases, and analyze the differences in the accuracy of each classifier.
For deep learning methods, this paper tests three repeatedly proven CNN methods and ViT, which have only recently been used in the field of image classification.
This paper focuses on the analysis of the four models from the accuracy rate, model size, training time and other indicators.
In addition, this paper conducts additional experiments for training time to find the best classification performance of ViT on GasHisSDB.
GasHisSDB demonstrates that the performance in the image classification experiments in this paper is competent to test existing image classification methods.

The creation of the dataset means that more image classification methods can be used for this dataset. We will try newer image classification methods on the GasHisSDB to compare and analyze image classification methods and obtain better practical methods to contribute to medical progress.

\section{Acknowledgements}
This work is supported by the "National Natural Science Foundation of China" (No. 61806047) and the "Fundamental Research Funds for the Central Universities" (No. N2019003). 
We also thank Miss. Zixian Li and Mr. Guoxian Li for their important discussion in this work.

\bibliographystyle{unsrt}
\bibliography{cas-refs}

\begin{thebibliography}{10}

\bibitem{Wang-2019-tcsoc}
Feng~Hua Wang, L.~Shen, J.~Li, Zhi~Wei Zhou, H.~Liang, Xiao~Tian Zhang,
  L.~Tang, Y.~Xin, J.~Jin, and Yu~Jing Zhang.
\newblock The chinese society of clinical oncology (csco): Clinical guidelines
  for the diagnosis and treatment of gastric cancer.
\newblock {\em Cancer Communications}, 39(1), 2019.

\bibitem{Sung-2021-gcs2g}
Hyuna Sung, Jacques Ferlay, Rebecca~L. Siegel, Mathieu Laversanne, Isabelle
  Soerjomataram, Ahmedin Jemal, and Freddie Bray.
\newblock {Global cancer statistics 2020: GLOBOCAN estimates of incidence and
  mortality worldwide for 36 cancers in 185 countries}.
\newblock {\em CA: A Cancer Journal for Clinicians}, 0(0):1--41, 2021.

\bibitem{Tom-2014-gceee}
Tom, Waddell, Marcel, Verheij, William, Allum, David, Cunningham, Andrés, and
  Cervantes.
\newblock Gastric cancer†: Esmo–esso–estro clinical practice guidelines
  for diagnosis, treatment and follow-up.
\newblock {\em Radiotherapy and Oncology}, 2014.

\bibitem{Fischer-2008-haeso}
Jack Rose Rolf~Zeller Andrew H~Fischer, Kenneth A~Jacobson.
\newblock Hematoxylin and eosin staining of tissue and cell sections.
\newblock {\em CSH protocols}, 2008.

\bibitem{Chan-2014-twcot}
John K~C Chan.
\newblock The wonderful colors of the hematoxylin-eosin stain in diagnostic
  surgical pathology.
\newblock {\em Int J Surg Pathol}, 22(1):12--32, 2014.

\bibitem{Cardiff-2014-mhaes}
Robert~D Cardiff, Claramae~H Miller, and Robert~J Munn.
\newblock Manual hematoxylin and eosin staining of mouse tissue sections.
\newblock {\em Cold Spring Harbor protocols}, 2014(6):655–658, 2014.

\bibitem{kumar-2017-rbpeb}
Vinay Kumar, Abul~K Abbas, and Jon~C Aster.
\newblock {\em Robbins basic pathology e-book}.
\newblock Elsevier Health Sciences, 2017.

\bibitem{wang-2021-miras}
Jing Wang and Xiuping Liu.
\newblock Medical image recognition and segmentation of pathological slices of
  gastric cancer based on deeplab v3+ neural network.
\newblock {\em Computer Methods and Programs in Biomedicine}, page 106210,
  2021.

\bibitem{tsekrekos-2019-htrgi}
Andrianos Tsekrekos, S{\"o}nke Detlefsen, Robert Riddell, James Conner, Luca
  Mastracci, Kieran Sheahan, Jayant Shetye, Lars Lundell, and Michael Vieth.
\newblock Histopathologic tumor regression grading in patients with gastric
  carcinoma submitted to neoadjuvant treatment: results of a delphi survey.
\newblock {\em Human pathology}, 84:26--34, 2019.

\bibitem{jin-2020-aiigc}
Peng Jin, Xiaoyan Ji, Wenzhe Kang, Yang Li, Hao Liu, Fuhai Ma, Shuai Ma, Haitao
  Hu, Weikun Li, and Yantao Tian.
\newblock Artificial intelligence in gastric cancer: A systematic review.
\newblock {\em Journal of Cancer Research and Clinical Oncology}, pages 1--12,
  2020.

\bibitem{mirniaharikandehei-2021-aarpa}
Seyedehnafiseh Mirniaharikandehei, Morteza Heidari, Gopichandh Danala,
  Sivaramakrishnan Lakshmivarahan, and Bin Zheng.
\newblock Applying a random projection algorithm to optimize machine learning
  model for predicting peritoneal metastasis in gastric cancer patients using
  ct images.
\newblock {\em Computer Methods and Programs in Biomedicine}, 200:105937, 2021.

\bibitem{le-2021-jcaol}
Trinh~Thi Le~Vuong, Kyungeun Kim, Boram Song, and Jin~Tae Kwak.
\newblock Joint categorical and ordinal learning for cancer grading in
  pathology images.
\newblock {\em Medical image analysis}, 73:102206, 2021.

\bibitem{Miranda-2016-asomi}
E.~{Miranda}, M.~{Aryuni}, and E.~{Irwansyah}.
\newblock A survey of medical image classification techniques.
\newblock In {\em 2016 International Conference on Information Management and
  Technology (ICIMTech)}, pages 56--61, 2016.

\bibitem{affonso-2017-dlfbi}
Carlos Affonso, Andr{\'e} Luis~Debiaso Rossi, F{\'a}bio Henrique~Antunes
  Vieira, Andr{\'e} Carlos~Ponce de~Leon~Ferreira, et~al.
\newblock Deep learning for biological image classification.
\newblock {\em Expert Systems with Applications}, 85:114--122, 2017.

\bibitem{kotadiya-2019-romic}
Hiral Kotadiya and Darshana Patel.
\newblock Review of medical image classification techniques.
\newblock In {\em Third International Congress on Information and Communication
  Technology}, pages 361--369. Springer, 2019.

\bibitem{sharma-2015-amafc}
Harshita Sharma, Norman Zerbe, Daniel Heim, Stephan Wienert, Hans-Michael
  Behrens, Olaf Hellwich, and Peter Hufnagl.
\newblock A multi-resolution approach for combining visual information using
  nuclei segmentation and classification in histopathological images.
\newblock In {\em VISAPP (3)}, pages 37--46, 2015.

\bibitem{behrens-2015-rohsi}
Hans-Michael Behrens, Viktoria~S Warneke, Christine B{\"o}ger, Nele Garbrecht,
  Eva J{\"u}ttner, Wolfram Klapper, Micaela Mathiak, Ilske Oschlies, Ursula
  Rudolph, Christiane Stuhlmann-Laeisz, et~al.
\newblock Reproducibility of her2/neu scoring in gastric cancer and assessment
  of the 10\% cut-off rule.
\newblock {\em Cancer medicine}, 4(2):235--244, 2015.

\bibitem{sharma-2017-acsof}
Harshita Sharma, Norman Zerbe, Christine B{\"o}ger, Stephan Wienert, Olaf
  Hellwich, and Peter Hufnagl.
\newblock A comparative study of cell nuclei attributed relational graphs for
  knowledge description and categorization in histopathological gastric cancer
  whole slide images.
\newblock In {\em 2017 IEEE 30th International Symposium on Computer-Based
  Medical Systems (CBMS)}, pages 61--66. IEEE, 2017.

\bibitem{TP-2017-psioa}
Pathological section identification of ai challenge.
\newblock \url{http://www.datadreams.org/race-data-3.html}, 2017.

\bibitem{liu-2018-cogsb}
Bo~Liu, Ming Zhang, Tongyu Guo, and Yuanzhi Cheng.
\newblock Classification of gastric slices based on deep learning and sparse
  representation.
\newblock In {\em 2018 Chinese Control And Decision Conference (CCDC)}, pages
  1825--1829. IEEE, 2018.

\bibitem{korkmaz-2018-comsi}
Sevcan~Ayta{\c{c}} Korkmaz and Hamidullah Binol.
\newblock Classification of molecular structure images by using ann, rf, lbp,
  hog, and size reduction methods for early stomach cancer detection.
\newblock {\em Journal of Molecular Structure}, 1156:255--263, 2018.

\bibitem{sharma-2017-dcnnf}
Harshita Sharma, Norman Zerbe, Iris Klempert, Olaf Hellwich, and Peter Hufnagl.
\newblock Deep convolutional neural networks for automatic classification of
  gastric carcinoma using whole slide images in digital histopathology.
\newblock {\em Computerized Medical Imaging and Graphics}, 61:2--13, 2017.

\bibitem{li-2018-dlbgc}
Yuexiang Li, Xuechen Li, Xinpeng Xie, and Linlin Shen.
\newblock Deep learning based gastric cancer identification.
\newblock In {\em 2018 IEEE 15th International Symposium on Biomedical Imaging
  (ISBI 2018)}, pages 182--185. IEEE, 2018.

\bibitem{wang-2019-rrmdl}
Shujun Wang, Yaxi Zhu, Lequan Yu, Hao Chen, Huangjing Lin, Xiangbo Wan, Xinjuan
  Fan, and Pheng-Ann Heng.
\newblock Rmdl: Recalibrated multi-instance deep learning for whole slide
  gastric image classification.
\newblock {\em Medical image analysis}, 58:101549, 2019.

\bibitem{zhu-2020-wbanf}
Zhonghang Zhu, Xin Ding, Defu Zhang, and Liansheng Wang.
\newblock Weakly-supervised balanced attention network for gastric pathology
  image localization and classification.
\newblock In {\em 2020 IEEE 17th International Symposium on Biomedical Imaging
  (ISBI)}, pages 1--4. IEEE, 2020.

\bibitem{song-2020-cahds}
Zhigang Song, Shuangmei Zou, Weixun Zhou, Yong Huang, Liwei Shao, Jing Yuan,
  Xiangnan Gou, Wei Jin, Zhanbo Wang, Xin Chen, et~al.
\newblock Clinically applicable histopathological diagnosis system for gastric
  cancer detection using deep learning.
\newblock {\em Nature communications}, 11(1):1--9, 2020.

\bibitem{kosaraju-2020-dmrfd}
Sai~Chandra Kosaraju, Jie Hao, Hyun~Min Koh, and Mingon Kang.
\newblock Deep-hipo: Multi-scale receptive field deep learning for
  histopathological image analysis.
\newblock {\em Methods}, 179:3--13, 2020.

\bibitem{sun-2020-ghisu}
Changhao Sun, Chen Li, Jinghua Zhang, Md~Mamunur Rahaman, Shiliang Ai, Hao
  Chen, Frank Kulwa, Yixin Li, Xiaoyan Li, and Tao Jiang.
\newblock Gastric histopathology image segmentation using a hierarchical
  conditional random field.
\newblock {\em Biocybernetics and Biomedical Engineering}, 40(4):1535--1555,
  2020.

\bibitem{zhang-2021-jfcag}
Jun Zhang, Zhiyuan Hua, Kezhou Yan, Kuan Tian, Jianhua Yao, Eryun Liu, Mingxia
  Liu, and Xiao Han.
\newblock Joint fully convolutional and graph convolutional networks for
  weakly-supervised segmentation of pathology images.
\newblock {\em Medical image analysis}, 73:102183, 2021.

\bibitem{li-2019-chicu}
Chen Li, Hao Chen, Le~Zhang, Ning Xu, Dan Xue, Zhijie Hu, He~Ma, and Hongzan
  Sun.
\newblock Cervical histopathology image classification using multilayer hidden
  conditional random fields and weakly supervised learning.
\newblock {\em Ieee Access}, 7:90378--90397, 2019.

\bibitem{mohamed-2021-fmrfa}
Mostafa~M Mohamed, Mina~A Nessiem, Anton Batliner, Christian Bergler, Simone
  Hantke, Maximilian Schmitt, Alice Baird, Adria Mallol-Ragolta, Vincent Karas,
  Shahin Amiriparian, et~al.
\newblock Face mask recognition from audio: The masc database and an overview
  on the mask challenge.
\newblock {\em Pattern Recognition}, page 108361, 2021.

\bibitem{potovcnik-2020-pdfvo}
Bo{\v{z}}idar Poto{\v{c}}nik, Jurij Munda, Milan Relji{\v{c}}, Ksenija
  Raki{\'c}, Jure Knez, Veljko Vlaisavljevi{\'c}, Ga{\v{s}}per Sedej, Boris
  Cigale, Ale{\v{s}} Holobar, and Damjan Zazula.
\newblock Public database for validation of follicle detection algorithms on 3d
  ultrasound images of ovaries.
\newblock {\em Computer Methods and Programs in Biomedicine}, 196:105621, 2020.

\bibitem{weiss-2007-eawst}
Sharon~W Weiss, John~R Goldblum, and Andrew~L Folpe.
\newblock {\em Enzinger and Weiss's soft tissue tumors}.
\newblock Elsevier Health Sciences, 2007.

\bibitem{japanese-2011-jcogc}
Japanese Gastric Cancer~Association jgca@ koto. kpu-m.~ac. jp.
\newblock Japanese classification of gastric carcinoma: 3rd english edition.
\newblock {\em Gastric cancer}, 14:101--112, 2011.

\bibitem{takahashi-2013-gccso}
Tsunehiro Takahashi, Yoshiro Saikawa, and Yuko Kitagawa.
\newblock Gastric cancer: current status of diagnosis and treatment.
\newblock {\em Cancers}, 5(1):48--63, 2013.

\bibitem{stevens-1990-th}
Alan Stevens.
\newblock The haematoxylins.
\newblock {\em Theory and practice of histological techniques}, 1990.

\bibitem{wittekind-2003-tsfrd}
D~Wittekind.
\newblock Traditional staining for routine diagnostic pathology including the
  role of tannic acid. 1. value and limitations of the hematoxylin-eosin stain.
\newblock {\em Biotechnic \& histochemistry}, 78(5):261--270, 2003.

\bibitem{gevers-2006-cfd}
Theo Gevers, Joost Van De~Weijer, and Harro Stokman.
\newblock Color feature detection, 2006.

\bibitem{van-2006-clfe}
Joost Van De~Weijer and Cordelia Schmid.
\newblock Coloring local feature extraction.
\newblock In {\em European conference on computer vision}, pages 334--348.
  Springer, 2006.

\bibitem{sharma-2001-eotmf}
Mona Sharma and Sameer Singh.
\newblock Evaluation of texture methods for image analysis.
\newblock In {\em The Seventh Australian and New Zealand Intelligent
  Information Systems Conference, 2001}, pages 117--121. IEEE, 2001.

\bibitem{zhou-2013-ltfbo}
Shu-Ren Zhou and Jian-Ping Yin.
\newblock Lbp texture feature based on haar characteristics.
\newblock {\em Journal of Software}, 24(8):1909--1926, 2013.

\bibitem{mohanaiah-2013-itfeu}
P~Mohanaiah, P~Sathyanarayana, and L~GuruKumar.
\newblock Image texture feature extraction using glcm approach.
\newblock {\em International journal of scientific and research publications},
  3(5):1--5, 2013.

\bibitem{qunqun-2013-eocit}
HOU Qunqun, WANG Fei, and YAN Li.
\newblock Extraction of color image texture feature based on gray-level
  co-occurrence matrix.
\newblock {\em Remote Sensing for Land \& Resources}, 25(4):26--32, 2013.

\bibitem{ali-2018-cagad}
Hussam Ali, Mussarat Yasmin, Muhammad Sharif, and Mubashir~Husain Rehmani.
\newblock Computer assisted gastric abnormalities detection using hybrid
  texture descriptors for chromoendoscopy images.
\newblock {\em Computer methods and programs in biomedicine}, 157:39--47, 2018.

\bibitem{liaw-2002-carbr}
Andy Liaw, Matthew Wiener, et~al.
\newblock Classification and regression by randomforest.
\newblock {\em R news}, 2(3):18--22, 2002.

\bibitem{hassoun-1995-foann}
Mohamad~H Hassoun et~al.
\newblock {\em Fundamentals of artificial neural networks}.
\newblock MIT press, 1995.

\bibitem{wang-2020-anact}
Liansheng Wang, Yudi Jiao, Ying Qiao, Nianyin Zeng, and Rongshan Yu.
\newblock A novel approach combined transfer learning and deep learning to
  predict tmb from histology image.
\newblock {\em Pattern Recognition Letters}, 135:244--248, 2020.

\bibitem{simonyan-2014-vdcnf}
Karen Simonyan and Andrew Zisserman.
\newblock Very deep convolutional networks for large-scale image recognition.
\newblock {\em arXiv preprint arXiv:1409.1556}, 2014.

\bibitem{he-2016-drlfi}
Kaiming He, Xiangyu Zhang, Shaoqing Ren, and Jian Sun.
\newblock Deep residual learning for image recognition.
\newblock In {\em Proceedings of the IEEE conference on computer vision and
  pattern recognition}, pages 770--778, 2016.

\bibitem{dosovitskiy-2020-aiiww}
Alexey Dosovitskiy, Lucas Beyer, Alexander Kolesnikov, Dirk Weissenborn,
  Xiaohua Zhai, Thomas Unterthiner, Mostafa Dehghani, Matthias Minderer, Georg
  Heigold, Sylvain Gelly, et~al.
\newblock An image is worth 16x16 words: Transformers for image recognition at
  scale.
\newblock {\em arXiv preprint arXiv:2010.11929}, 2020.

\end{thebibliography}

\end{document}